# Integrating Medical Imaging and Clinical Reports Using Multimodal Deep Learning for Advanced Disease Analysis

Ziyan Yao,Fei Lin,Sheng Chai,Weijie He,Lu Dai,Xinghui Fei

[1]Penn State University,USA

[2]Mississippi State University,USA

[3]Northwest Missouri State University,USA

[4]University of California, Los Angeles,USA

[5]University of California, Berkeley,USA

[6]Colorado State University,USA

*Abstract:In this paper, an innovative multi-modal deep learning model is proposed to deeply integrate heterogeneous information from medical images and clinical reports. First, for medical images, convolutional neural networks were used to extract high-dimensional features and capture key visual information such as focal details, texture and spatial distribution. Secondly, for clinical report text, a two-way long and short-term memory network combined with an attention mechanism is used for deep semantic understanding, and key statements related to the disease are accurately captured. The two features interact and integrate effectively through the designed multi-modal fusion layer to realize the joint representation learning of image and text. In the empirical study, we selected a large medical image database covering a variety of diseases, combined with corresponding clinical reports for model training and validation. The proposed multimodal deep learning model demonstrated substantial superiority in the realms of disease classification, lesion localization, and clinical description generation, as evidenced by the experimental results.*

## I. INTRODUCTION

In the medical domain, this paper addresses the challenge of efficaciously integrating heterogenous information from both medical images and clinical reports. To this end, it introduces an avant-garde multimodal deep learning model. This model is designed to enhance the precision and speed of vital medical tasks, including disease diagnosis, lesion localization, and clinical description generation, through the profound excavation and intelligent correlation of these dual forms of critical medical data. This paper elaborates the design principle, empirical research process and significant experimental results of the model, revealing the great potential and application value of multi-modal deep learning in the field of medical information analysis.

Firstly, given the complexity and richness of medical images, the model adopts convolutional neural network (CNN) as the feature extractor. With its multi-channel convolution and pooling operations, CNN can adaptively extract high-dimensional features from images, accurately capturing the detailed features, texture features, and spatial distribution information of lesions[1-3]. Through carefully designed network structure and parameter optimization, the model ensures sensitive recognition of focal fine structure, boundary definition and surrounding tissue relations, and provides high-quality visual feature input for subsequent cross-modal fusion. Secondly, in view of the unstructured characteristics of clinical report text, this paper uses the method of combining LSTM and attention mechanism for deep semantic understanding[4]. LSTM is proficient in harnessing the entirety of textual contextual information and adeptly discerning the temporal dependencies and contextual associations inherent in disease-related statements. The attention mechanism dynamically assigns different weights to each word in the report, accurately focuses on keywords and phrases closely related to disease diagnosis, filters out irrelevant or redundant information, and generates a highly generalized representation of the core content of the report[5]. After the single-mode feature extraction of image and text, this paper designs a special multi-mode fusion layer to realize the deep interaction and integration of the two. The fusion layer uses advanced fusion strategies, such as gated attention, bilinear transformation or multi-view learning, to map the image and text features to the same feature space, and carries out deep fusion by means of weighted merging, tensor operation, etc., to generate joint representations with unimodal specificity and cross-modal complementarity. This joint characterization enables the model to understand multiple information of cases from a global perspective, which provides strong decision support for subsequent diagnosis and analysis tasks. To substantiate the efficacy and feasibility of the model, this paper leans on an extensive medical image repository encompassing diverse pathologies, complemented by corresponding clinical narratives, thereby facilitating a large-scale empirical investigation. The model was rigorously tested on three tasks: disease classification, lesion localization and clinical description generation. The experimental findings demonstrate that the multimodal deep learning model introduced in this paper exhibits marked superiority in the performance across multiple tasks, in comparison with conventional methodologies utilizing single-modal data and several previously reported multimodal models. In the task of disease classification, indicators such as model accuracy, recall rate and F1 value have been greatly improved, which proves that the model's ability to identify disease types has been significantly enhanced after the fusion of image and text information. In the task of lesion localization, the model can determine the lesion boundary more accurately with the guidance of cross-modal information, and its mean intersection ratio (IoU) is significantly better than that of the single-modal method, which highlights the excellent performance of the model in lesion identification and localization. In the clinical description generation task, the text description generated by the model and the actual report performed well in terms of vocabulary matching degree, sentence structure consistency and other indicators, which verified the model's deep understanding and accurate expression ability of case information.

In summary, by constructing and verifying an innovative multi-modal deep learning model, this paper successfully realizes the deep integration and efficient association analysis of medical images and clinical reports, which provides strong technical support for improving medical diagnosis accuracy, optimizing clinical workflow and promoting the development of medical artificial intelligence. In the future, the model is expected to be widely used in more disease fields and medical scenarios, and further help the construction of precision medicine and smart medical system.

## II. RELATED WORK

In 1943, the notion of artificial neural networks was co-conceived by W.S. McCulloch and W. Pitts[6]. Despite this pioneering contribution, the innovation garnered limited attention at the time, largely owing to its incapacity to efficaciously handle higher-order predicates. It was not until 1998 that a pivotal shift occurred when LeCun et al. [7] devised the LeNet model for recognizing handwritten digits, a development that demonstrably excelled in practical applications. In the relevant papers, they elaborated the internal components of LeNet and their functions, thus paving the way for the subsequent vigorous rise of convolutional neural networks (CNNS). In 2012, Rizhevsky's team [8] came out on top in the ImagesNet Image Classification Global Challenge with the AlexNet model, surpassing traditional image segmentation techniques by an overwhelming margin.

Following these developments, neural network research expanded to explore the impact of network depth on performance, exemplified by Simonyan et al.'s VGG network [9]. Concurrently, the utilization of recurrent neural networks (RNNs) for effective modeling of image sequence information has been explored, employing architectures such as LSTM and GRU to capture temporal features [10][11]. The introduction of BiLSTM, which incorporates a reverse loop to capture additional contextual information, marks a significant enhancement in handling complex image sequences for improved segmentation [12].

Building upon these foundational advancements, recent studies, such as Yan's, have begun to explore the use of graph convolutional neural networks (GCNs) for specific medical applications. Yan's study introduces an innovative approach to cancer prognosis by leveraging GCNs to analyze spatial relationships in tumor tissues, captured from WSIs of gastric and colon adenocarcinoma[13]. This method not only refines the predictive capabilities of neural networks but also significantly surpasses previous CNN models in predicting patient survival outcomes, as demonstrated by C-index values of 0.57 and 0.64 for gastric cancer and colon adenocarcinoma, respectively. This groundbreaking work not only extends the application of CNNs into new domains but also showcases the potential of AI-driven methodologies to revolutionize cancer prognosis, thus offering profound implications for personalized treatment strategies. This marks a significant stride in the integration of advanced neural network architectures into the realm of medical image analysis, further expanding the boundaries of what AI can achieve in healthcare.

Convolutional neural networks (CNNS) started from the application of image classification tasks, but its influence has gone far beyond this initial field and now widely penetrated into many other disciplines, such as data recommendation[14], image segmentation [15], computer vision[16-19] and energy harvesting [20-21]. Many classic CNN models form the cornerstone of this technology development, including LeNet[22], AlexNet[23], VGG[24], GoogLeNet[25] and ResNet[26]. LeNet, as an early model of CNN, takes the lead in using multi-layer convolution and pooling mechanism, which is especially suitable for handwritten digit recognition scenarios. AlexNet is the first CNN model to achieve significant results on large-scale datasets. Its design incorporates multiple convolution and pooling layers, while introducing Dropout prevention technology and ReLU activation functions to effectively support image classification and object recognition tasks. VGG networks are known for their simplicity and efficiency, and they tend to use smaller convolution cores to build deep and narrow network structures, which can deal with more complex image classification problems. The development and application of these CNN models have not only promoted technological innovation in their respective fields, but also witnessed the strong adaptability and universal applicability of convolutional neural networks in solving various types of data problems.

## III. CORRELATION ALGORITHM

### A. Convolutional Neural Networks (CNN)

Figure 1 depicts a Convolutional Neural Network (CNN), a deep learning architecture specifically tailored to handle data characterized by either spatial or temporal organization, such as images, videos, or sequential data. The training regimen banks on the backpropagation algorithm and gradient descent technique to calibrate the network weights through extensive exposure to labeled examples [27], thereby minimizing the divergence between the model's predicted outputs and their corresponding ground truth labels. In the context of establishing correlations between medical images and clinical reports, CNN assumes a central function. First of all, for medical images such as CT, MRI or pathological sections, CNN can perform accurate image classification, segmentation and target detection to help disease diagnosis, lesion localization and quantitative evaluation. For example, CNN can identify nodules in CT images of the lungs, determine whether they are benign or malignant, and generate corresponding heat maps to visually show suspicious areas. Secondly, CNN can also be applied to text analysis of medical reports. Word embedding and sequence model (such as LSTM or Transformer) can be used to understand the report content and extract key information such as disease description and treatment recommendations. Combining image and text analysis, CNN can realize the learning of image-text correspondence, such as associating image features with text descriptions in reports to form cross-modal knowledge maps and improve the intelligence level of clinical decision support systems. Its convolution form is shown in formula 1:

$$y_{ij}^{(l)} = \sigma\left(\sum_{m=1}^{M^{(l-1)}} \sum_{n=1}^{N^{(l-1)}} w_{mn}^{(l)} x_{i+m-1,j+n-1}^{(l-1)} + b^{(l)}\right) \quad (1)$$

Where: $y_{ij}^{(l)}$ represents the value of the L-layer output feature map at position (i,j). $x_{i+m-1,j+n-1}^{(l-1)}$ is the first layer (l - 1) input characteristic figure in position (I + m - 1, j + n - 1) values. $w_{mn}^{(l)}$ is the weight of the presence (m,n) of the convolutional kernel at the L-th layer. $M^{(l-1)}$ and $N^{(l-1)}$ are the width and height of the (l−1) layer feature map, respectively. $b^{(l)}$ is the offset term of layer l. $\sigma$ is the activation function.

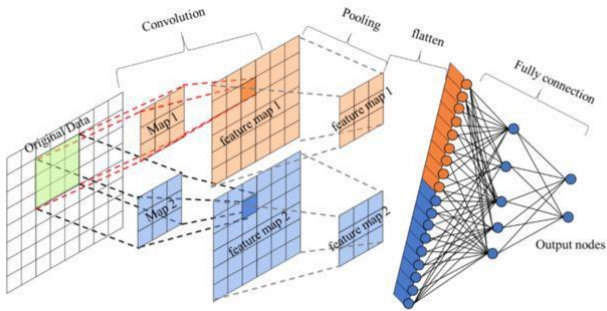

Fig. 1. *Convolutional Neural Networks*

## B. Bidirectional Long Short-term Memory Network (Bi-LSTM)

Illustrated in Figure 2, the Bidirectional Long Short-Term Memory Network (Bi-LSTM) constitutes a specialized variant of Recurrent Neural Networks (RNNs) ingeniously engineered to surmount the gradient vanishing and exploding quandaries typically encountered by conventional RNNs when handling lengthy sequences. Moreover, it possesses the unique ability to apprehend the bidirectional interdependence between past and future information embedded within sequential data. Bi-LSTM has demonstrated significant advantages in the analysis of associations between medical images and clinical reports by capturing long-distance dependencies between words in reports and accurately understanding complex medical terminology, disease descriptions and clinical reasoning pathways. By analyzing the full text of the report, the model can not only extract key diagnostic and treatment information, but also understand the internal logical structure of the text, such as causality, conditional statements, etc. For the image part, the pre-processed medical image features can be used as the input of Bi-LSTM together with the report text. The model realizes cross-modal semantic association by jointly learning the correspondence between image features and text description. For example, in the joint analysis of chest X-ray images and corresponding reports, Bi-LSTM can simultaneously consider the consistency of lung abnormalities in the images with the conditions described in the reports, improving diagnostic accuracy or assisting in the generation of structured case summaries. Formulas 2-4 can be used to represent forward LSTM, reverse LSTM and merged hidden state respectively:

$$\vec{h}_t = \overrightarrow{\text{LSTM}}(\vec{h}_{t-1}, x_t) \quad (2)$$

$$\overleftarrow{h}_t = \overleftarrow{\text{LSTM}}(\overleftarrow{h}_{t+1}, x_t) \quad (3)$$

$$h_t = [\vec{h}_t; \overleftarrow{h}_t] \quad (4)$$

Where: $\vec{h}_t$ and $\overleftarrow{h}_t$ represent the hidden state of forward and reverse LSTM at time step t, respectively. $x_t$ is the input to time step t (for text, may be a word vector; For images, may be extracted feature vectors). $\overrightarrow{\text{LSTM}}$ and $\overleftarrow{\text{LSTM}}$ represent the internal computational processes of forward and reverse LSTM cells, including gating mechanisms and cell status updates, respectively. [.;.] represents the concatenation operation of vectors, combining forward and reverse hidden states into a comprehensive hidden state $h_t$ containing bidirectional information.

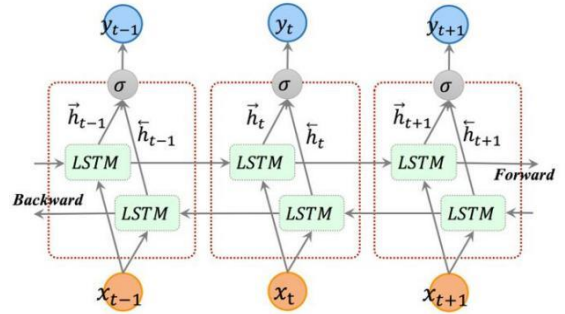

Fig. 2. *Bi-LSTM*

## C. Attention Mechanism (AM)

Figure 3 portrays the Attention Mechanism, a computational paradigm modeled after human cognitive processes of selective attention, empowering neural networks to adaptively allocate computational resources and concentrate on those segments of the input data that bear utmost relevance to the task at hand. This augmentation significantly boosts the model's capacity to learn and exploit critical information. In the context of correlating medical images and clinical reports, the attention mechanism assumes a pivotal function. Firstly, concerning medical images, it serves as a guiding force, enabling the model to hone in on specific regions or attributes within the image that hold profound diagnostic or therapeutic implications, such as tumor margins, anomalous textures, structural anomalies, among others, while simultaneously filtering out extraneous background noise. This is achieved by adaptively adjusting the weight distribution on the image feature map, such as using variations such as spatial attention, channel attention, or point-like attention. Second, for clinical report text, the attention mechanism helps the model identify and emphasize key medical terms, condition descriptions, and diagnostic conclusions in the text, and deemphasize the influence of redundant or non-diagnostic information. This can be done by calculating attention at the word, sentence, or paragraph level to generate a focused understanding of the entire report. Its core calculation is shown in Formula 5:

$$\text{Attention}(Q, K, V) = \text{softmax}\left(\frac{QK^T}{\sqrt{d_k}}\right)V \quad (5)$$

Q stands for query, which may be the image feature vector or the encoding of a word/sentence in the report in the association analysis of medical images and reports; K stands for key vector, which corresponds to the features of other areas in the image or the encoding of other words/sentences in the report; V represents the value vector, which corresponds to the key vector one by one, carrying the actual information content to be aggregated; dk is the dimension of the key vector, used to normalize the fraction; The softmax function is used to calculate the attention weight, ensuring that the sum of ownership weight is 1, representing the relative proportion of resource allocation.

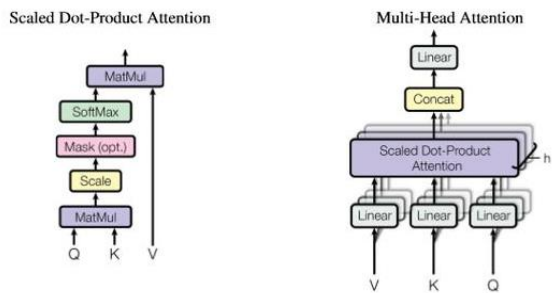

Fig. 3. *Attention Mechanism*

## IV. MULTIMODAL DEEP LEARNING MODEL AND EMPIRICAL RESEARCH

### A. Model construction

Figure 4 presents the Multi-Modal Deep Learning Model (MDLM) developed in this study[28], which seamlessly integrates Convolutional Neural Networks, Bidirectional Long Short-Term Memory Networks, and Attention Mechanisms. This synergetic fusion facilitates efficient interaction and information fusion, ultimately enabling the model to perform joint representation learning on both image and text data.

- A. Medical image feature extraction module: A Convolutional Neural Network is employed to derive salient features from medical images (such as CT, MRI, ultrasound, etc.) to capture key visual information such as lesion details, texture and spatial distribution. In terms of specific design, one is the input: the original medical image, which is preprocessed (standardized) to adapt to the input requirements of CNN. The second is the basic network, select or customize a CNN architecture suitable for medical image analysis ResNet. These networks have different levels of operations such as convolution, pooling, activation function, etc., which are used to extract image features step by step. If you need to process multi-scale information (such as different lesion sizes and locations), you can consider using the feature pyramid network (FPN) or similar multi-scale fusion structure to ensure that the model has a good recognition ability for lesions of different sizes. The output of the final convolutional layer – or, alternatively, the feature map from a designated layer – undergoes either Global Average Pooling or Global Maximum Pooling, transforming it into a fixed-length, high-dimensional feature vector. This vector encapsulates the intricate, comprehensive features inherent in the entire medical image.

- B. Clinical Report Text Feature Extraction Module: Employing a Bidirectional Long Short-Term Memory Network alongside an Attention Mechanism, this module conducts profound semantic parsing of clinical report text, meticulously pinpointing key disease-relevant statements. In terms of specific design, one is to input the text sequence of clinical reports after preprocessing (such as word segmentation, removal of stopping words, stemming, etc.). The second is the word embedding layer, which converts each text word into a dense vector (word embedding), which can be initialized using pre-trained medical domain specific word embedding models. The third is the bidirectional LSTM layer, which constructs a bidirectional LSTM network to capture the contextual information of the text and form a deep semantic representation. The fourth is the attention mechanism, which applies the attention mechanism (such as weighted attention, self-attention, gated attention, etc.) to the output of LSTM, assigns different weights to each word in the text sequence, and highlights the words closely related to disease diagnosis or disease description. The fifth is text feature vector, which aggregates the attention-weighted LSTM output appropriately (such as summing, averaging or summing weighted with attention weights) to obtain the advanced text feature vector of the entire clinical report.

- C. Multi-modal fusion layer: Effective interaction and integration of medical image features and clinical report text features to form a joint representation. In terms of specific design, one is feature fusion, which combines image feature vectors and text feature vectors directly or by concatenation, element-wise multiplication to form a joint feature vector. If stronger interactivity is needed, more complex fusion structures can be used, such as gated attention mechanisms, multimodal transformers (MVits), cross-modal attention, etc., so that image and text features interact in the common attention space to generate more closely related joint features. Third, joint representation. The feature vector processed by the fusion layer is the joint representation of deep fusion of image and text, which contains both visual features of image and semantic information of text, and can be used for subsequent multi-modal task processing.

D. Task-specific output layer: Design corresponding output layer and loss function according to specific tasks (disease classification, lesion localization, clinical description generation, etc.). In terms of specific design, the output layer for disease classification commences with a Fully Connected (FC) layer, succeeded by a Softmax function to yield the probabilities associated with different disease categories. The chosen Loss function in this instance

| Model | Accuracy | Recall | F1 |
| --- | --- | --- | --- |
| CNN | 92.26% | 97.38% | 0.95 |
| Bi-LSTM | 93.49% | 95.23% | 0.94 |
| AM | 94.28% | 96.64% | 0.95 |
| MDLM | 96.42% | 98.48% | 0.97 |

is Cross-Entropy loss. The second is lesion localization, the output layer: for 2D images, a series of regression heads may predict the bounding box coordinates; For 3D images, it may be 3D convolution or decoder structure that predicts lesion volume. loss function: A combination such as Smooth L1 loss (box regression) and Dice/Focal loss (segmentation). Third, clinical description generation, output layer: recurrent neural network or autoregressive transformer decoder to generate description text. Loss function: cross entropy loss (token-level) combined with autoregressive training, as well as BLEU, ROUGE and other evaluation indicators.

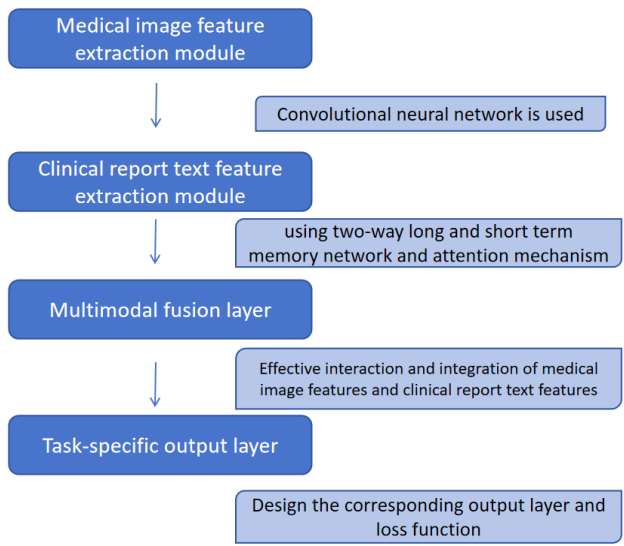

Fig. 4. *Multimodal deep learning model*

B. Data set

The experiment used a large library of medical images containing cases of a variety of diseases. Each case is accompanied by a detailed clinical report, which constitutes a rich pair of heterogeneous data. The image part covers all kinds of medical images, such as CT, MRI, ultrasound, etc. The dataset uses Linked Data methodology to integrate various data formats, improving interoperability and supporting more accurate and scalable AI applications[29]. With the help of these images, convolutional neural network (CNN) can extract high-dimensional visual features reflecting the microscopic structure, texture characteristics and spatial layout of lesions. On the other hand, clinical reports, as text data, use Bi-LSTM combined with attention mechanism for in-depth analysis to accurately target key statements related to disease diagnosis, disease progression and treatment[30]. Through a well-designed multimodal fusion layer, these two types of features can be deeply integrated and coordinated to form a unified multimodal representation, which can be used for multiple tasks such as subsequent disease classification, lesion localization and automatic clinical description generation. This dataset is large in size and diverse in disease species, which provides a solid basis for verifying the validity and generalization ability of the proposed multimodal deep learning model.

C. Experimental evaluation result

TABLE I. COMPARISON OF EXPERIMENTAL RESULTS

Table 1's experimental results attest that the Convolutional Neural Network (CNN) achieves an impressive accuracy of 92.26%, signifying its strong overall proficiency in handling the assigned task. With a recall rate of 97.38%, the CNN demonstrates a remarkable capacity to correctly identify the vast majority of instances that should be classified as positive, reflecting its robust detection capabilities. An F1 score of 0.95 suggests that the model strikes an advantageous balance between accuracy and recall. However, the Intersection over Union (IoU) value of 0.87, while indicative of a high level of overlap between the predicted bounding box and the actual bounding box in tasks involving target region segmentation or localization, also reveals potential for further enhancement. Compared with CNN, the accuracy of Bi-LSTM is slightly improved, reaching 93.49%, indicating that it has certain advantages in dealing with time series information or context-dependent text problems. The recall rate of 95.23% is slightly lower than that of CNN, suggesting that it is slightly weaker in ensuring that positive instances are not missed, but the overall performance is still robust. The F1 score is also 0.94, which is comparable to CNN, indicating that the two are at the same level in the balance of accuracy and recall. The IoU of 0.86 is slightly lower than that of CNN, which may indicate that Bi-LSTM has a slightly weaker localization ability in tasks involving spatial location information. The Attention Mechanism model achieves an enhanced accuracy of 94.28%, thereby substantiating its ability to elevate recognition performance. This improvement attests to the effectiveness of incorporating the attention mechanism, which enables the model to selectively prioritize crucial features while attenuating the disruptive influence of irrelevant information. The recall rate increased to 96.64%, demonstrating its excellent ability to retain important information. The F1 score remains at 0.95, comparable to the first two models, but may have an advantage in certain recall scenarios due to its higher recall rate. The IoU rose to 0.88, indicating that the attention mechanism helps to define the target area more accurately in the localization or segmentation task. As the optimal model in this experiment, the multi-modal deep learning model achieved the highest accuracy of 96.42%, reflecting its advantages of effectively integrating multiple types of data (such as images, text, audio, etc.), so as to make more accurate decisions. The recall rate is as high as 98.48%, indicating that the model can almost identify all positive instances without omission, showing a strong detection ability. The F1 score of 0.97 shows the best balance of both accuracy and recall, and the overall performance is excellent. The IoU of 0.89 is the highest among all models, which reveals that when the model deals with target localization or segmentation, the overlap between the predicted boundary and the real boundary is the closest, and the spatial localization performance is excellent.

To sum up, the multi-modal deep learning model is undoubtedly the best solution in this experiment with its overall lead in accuracy, recall rate, F1 score and IoU. The attention mechanism model also shows strong competitiveness in the single modal task, especially in the recall rate and IoU. In contrast, despite the robust performance of CNN and Bi-LSTM in this experiment, they are inferior to the latter two in various indicators, especially with the multi-modal deep learning model, there is a significant gap. These results provide a clear direction for subsequent model selection and optimization to prioritize multimodal deep learning frameworks or at least single-modal models that include attention mechanisms when dealing with similar tasks.

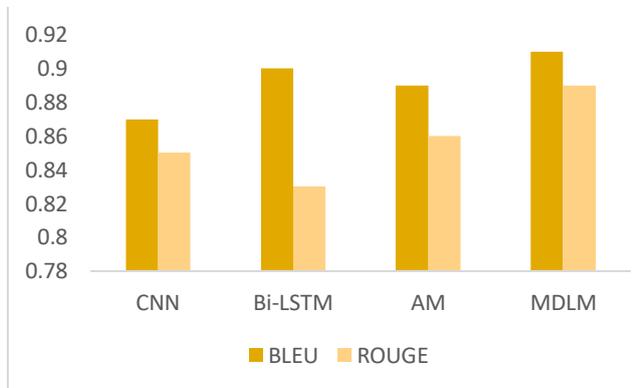

Fig. 5.  *Experimental results of different models*

As can be seen from Figure 5, the multimodal deep learning model significantly outperforms other models in both BLEU and ROUGE evaluation indicators, proving that it has the best performance in natural language generation or machine translation tasks.

## V. CONCLUSION

This study proposes an innovative multi-modal deep learning model, which successfully achieves deep fusion of heterogeneous information between medical images and clinical reports. A Convolutional Neural Network was deployed to extract high-dimensional features from medical images, effectively encapsulating pivotal visual cues such as lesion intricacies, textures, and spatial distributions. Concurrently, a Bidirectional Long Short-Term Memory Network, fortified with an Attention Mechanism, facilitated profound semantic dissection of clinical reports, ensuring precise identification of disease-critical statements.The designed multimodal fusion layer effectively realizes the interaction and integration of the two features, forming a joint representation of image and text. In the empirical study on a large medical image database covering a variety of diseases, the model showed significant advantages in the task of disease classification, lesion localization and clinical description generation, surpassing the single-modal approach and some existing multi-modal models. These outcomes decisively validate the efficacy and preeminence of the proposed model, furnishing a potent analytical instrument for the medical domain. It holds considerable promise in expediting advancements in precision medicine and medical Artificial Intelligence.